\documentclass[runningheads]{llncs}

\usepackage{subfigure}
\usepackage{graphicx}
\usepackage{amsmath}
\usepackage{bm}
\usepackage{url}
\usepackage{stmaryrd}
\usepackage{balance}
\usepackage{amssymb}
\usepackage{pgfplots}
\usepackage{pifont}
\usepackage{epsfig}
\usepackage{booktabs}
\usepackage{pgffor}
\usepackage{tikz}
\usepackage{multirow}
\usepackage{mathrsfs}
\usepackage{bbm}
\usepackage{helvet}
\usepackage{courier}
\usepackage{threeparttable}
\usepackage{algpseudocode} 
\usepackage{arydshln}
\usepackage{mathrsfs}
\usepackage{rotating} 
\usepackage{adjustbox}

\newcommand{\repeatthanks}{\textsuperscript{\thefootnote}}

\newcommand{\mb}{\mathbf}

\newcommand{\concatenate}{\operatornamewithlimits{\|}}
\newcommand{\our}{\text{LCGNN}}

\title{Label Contrastive Coding based Graph Neural Network for Graph Classification }
\author{Yuxiang Ren\inst{1}\thanks{Two authors contributed equally to this work} \and
	Jiyang Bai\inst{2}\repeatthanks \and
	Jiawei Zhang\inst{1}}
\authorrunning{Y. Ren et al.}
\institute{IFM Lab, Department of Computer Science, Florida State University, FL, USA\\
	\email{yuxiang@ifmlab.org, jiawei@ifmlab.org} \and
	Department of Computer Science, Florida State University, FL, USA
	\email{bai@cs.fsu.edu}}

\begin{document}
\titlerunning{ }
\maketitle

\begin{abstract}

Graph classification is a critical research problem in many applications from different domains. In order to learn a graph classification model, the most widely used supervision component is an output layer together with classification loss (e.g.,cross-entropy loss together with softmax or margin loss). In fact, the discriminative information among instances are more fine-grained, which can benefit graph classification tasks. In this paper, we propose the novel \textbf{L}abel \textbf{C}ontrastive Coding based \textbf{G}raph \textbf{N}eural \textbf{N}etwork ({\our}) to utilize label information more effectively and comprehensively. {\our} still uses the classification loss to ensure the discriminability of classes. Meanwhile, {\our} leverages the proposed \textit{Label Contrastive Loss} derived from self-supervised learning to encourage instance-level intra-class compactness and inter-class separability. To power the contrastive learning, {\our} introduces a dynamic label memory bank and a momentum updated encoder. Our extensive evaluations with eight benchmark graph datasets demonstrate that {\our} can outperform state-of-the-art graph classification models. Experimental results also verify that {\our} can achieve competitive performance with less training data because {\our} exploits label information comprehensively.
\end{abstract}

\section{Introduction}\label{sec:introduction}
Applications in many domains in the real world exhibit the favorable property of graph data structure, such as social networks~\cite{meng2019deep}, financial platforms~\cite{ren2019ensemfdet} and bioinformatics~\cite{FBSB17}. Graph classification aims to identify the class labels of graphs in the dataset, which is an important problem for numerous applications. For instance, in biology, a protein can be represented with a graph where each amino acid residue is a node, and the spatial relationships between residues (distances, angles) are the edges of a graph. Classification of graphs representing proteins can help predict protein interfaces~\cite{FBSB17}.

Recently, graph neural networks (GNNs) have achieved outstanding performance on graph classification tasks~\cite{xu2018powerful,zhang2019hierarchical}. GNNs aims to transform nodes to low-dimensional dense embeddings that preserve graph structural information and attributes~\cite{zhu2020deep}. When applying GNNs to graph classification, the standard method is to generate embeddings for all nodes in the graph and then summarize all these node embeddings to a representation of the entire graph, such as using a simple summation or neural network running on the set of node embeddings~\cite{ying2018hierarchical}. For the representation of the entire graph, a supervision component is usually utilized to achieve the purpose of graph classification. A final output layer together with classification loss (e.g.,cross-entropy loss together with softmax or margin loss) is the most commonly used supervision component in many existing GNNs~\cite{xu2018powerful,xinyi2018capsule,zhang2018end,hamilton2017inductive}. This supervision component focuses on the discriminability of class but ignores the instance-level discriminative representations. A recent trend towards learning stronger representations to serve classification tasks is to reinforce the model with discriminative information as more as possible~\cite{elsayed2018large}. To be explicit, graph representations, which consider both intra-class compactness and inter-class separability~\cite{liu2016large}, are more potent on the graph classification tasks. 

Inspired by the idea of recent self-supervised learning~\cite{chen2020simple} and contrastive learning~\cite{he2020momentum,qiu2020gcc}, the contrastive loss~\cite{oord2018representation} is able to extract extra discriminative information to improve the model's performance. The recent works~\cite{hjelm2018learning,qiu2020gcc,zhuang2019local} of using contrast loss for representation learning are mainly carried out under the setting of unsupervised learning. These contrastive learning models treat each instance as a distinct class of its own. Meanwhile, discriminating these instances is their learning objective~\cite{he2020momentum}. The series of contrastive learning have been verified effective in learning more fine-grained instance-level features in the computer vision~\cite{wu2018unsupervised} domain. Thus we plan to utilize the contrastive learning on graph classification tasks to make up for the shortcomings of supervision components, that is, ignoring the discriminative information on the instance-level. However, when applying contrastive learning, the inherent large intra-class variations may import noise to graph classification tasks~\cite{liu2016large}. Besides, existing contrastive learning based GNNs (e.g., GCC~\cite{qiu2020gcc}) detach the model pre-training and fine-tuning steps. Compared with end-to-end GNNs, the learned graph representations via contrastive learning can hardly be used in the downstream application tasks directly, like graph classification.


To cope with the task of graph classification, we propose the label contrastive coding based graph neural network ({\our}), which employs \textit{Label Contrastive Loss} to encourage instance-level intra-class compactness and inter-class separability simultaneously. Unlike existing contrastive learning using a single positive instance, the label contrastive coding imports label information and treats instances with the same label as multiple positive instances. In this way, the instances with the same label can be pulled closer, while the instances with different labels will be pushed away from each other.
Intra-class compactness and inter-class separability are taken into consideration simultaneously. The label contrastive coding can be regarded as training an encoder for a dictionary look-up task~\cite{he2020momentum}. In order to build an extensive and consistent dictionary, we propose a dynamic label memory bank and a momentum-updated graph encoder inspired by the mechanism~\cite{he2020momentum}. At the same time, {\our} also uses \textit{Classification Loss} to ensure the discriminability of classes. {\our} can utilize label information more effectively and comprehensively from instance-level and class-level, allowing using fewer label data to achieve comparative performance, which can be considered as a kind of label augmentation in essence. We validate the performance of {\our} on graph classification tasks over eight benchmark graph datasets. {\our} achieves SOTA performance in seven of the graph datasets. What is more, {\our} outperforms the baseline methods when using less training data, which verifies its ability to learn from label information more comprehensively. 

The contributions of our work are summarized as follows:
\begin{itemize}
	\item We propose a novel label contrastive coding based graph neural network ({\our}) to reinforce supervised GNNs with more discriminative information. 
	\item The \textit{Label Contrastive Loss} extends the contrastive learning to the supervised setting, where the label information can be imported to ensure intra-class compactness and inter-class separability.  
	\item The momentum-updated graph encoder and the dynamic label memory bank are proposed to support our supervised contrastive learning. 
	\item
	We conduct extensive experiments on eight benchmark graph datasets. {\our} not only achieves SOTA performance on multiple datasets but also can offer comparable results with fewer labeled training data.
\end{itemize}


\begin{figure}[t]
	\centering
	\vspace{-60pt}
	\begin{minipage}[l]{1\columnwidth}
		\centering
		\includegraphics[width=1.1\textwidth]{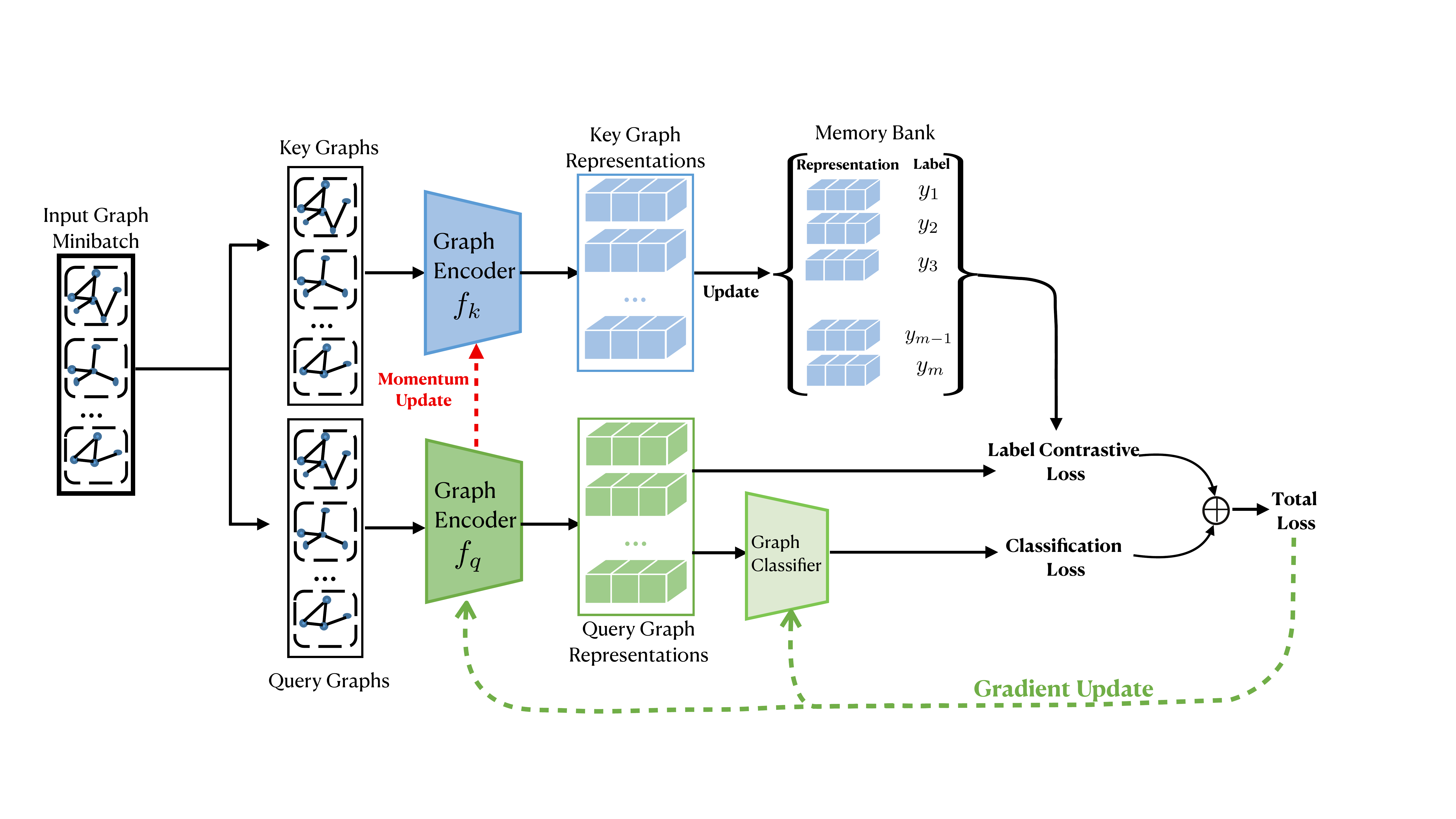}
	\end{minipage}
	\vspace{-40pt}
	
	\caption{\small The high-level structure of {\our}. {\our} trains the graph encoder $f_q$ and the graph classifier using a mixed loss. \textit{Label Contrastive Loss} and \textit{Classification Loss} constitute the mixed loss. \textit{Classification Loss} used in {\our} is cross-entropy loss. \textit{Label Contrastive Loss} is calculated by a dictionary look-up task. The query is each graph of the input graph minibatch, and the dictionary is a memory bank that can continuously update the label-known graph representations. The graph representation in the memory bank is updated by the graph encoder $f_k$, which is momentum-updated. After training, the learned graph encoder $f_q$, and the graph classifier can serve for graph classification tasks.}
	\label{fig:framework}
	
	\vspace{-10pt}
\end{figure}

\vspace{-15pt}
\section{Related Works} \label{sec:relatedwork}
\textbf{Graph Classification} \ Several different techniques have been proposed to solve the graph classification problem. One important category is the kernel-based method, which learns a graph kernel to measure similarity among graphs to differentiate graph labels~\cite{vishwanathan2010graph}. The Weisfeiler-Lehman subtree kernel (WL)~\cite{shervashidze2011weisfeiler}, Multiscale Laplacian graph kernels (MLG)~\cite{kondor2016multiscale}, and Graphlets kernel(GK)~\cite{shervashidze2009efficient} are all representative graph kernels. Another critical category is the deep-learning-based method. Deep Graph Kernel (DGK)~\cite{yanardag2015deep}, Anonymous Walk Embeddings (AWE), and  Graph2vec~\cite{narayanan2017graph2vec} all employ the deep-learning framework to extract the graph embeddings for graph classification tasks. With the rise of graph neural networks (GNNs), many GNNs are also used for graph classification tasks by learning the representation of graphs, which will be introduced below.\\   
\textbf{Graph Neural Network} \ The graph neural network learns the low-dimensional graph representations through a recursive neighborhood aggregation scheme~\cite{xu2018powerful}. The derived graph representations can be used to serve various downstream tasks, such as graph classification and top-k similarity search. According to the learning method, the current GNN that can serve graph classification can be divided into end-to-end models and pre-train models. The end-to-end models are usually under supervised or semi-supervised settings, with the goal of optimizing classification loss or mutual information, mainly including GIN~\cite{xu2018powerful}, CapsGNN~\cite{xinyi2018capsule}, DGCNN~\cite{zhang2018end} and InfoGraph~\cite{sun2019infograph}.The pre-trained GNNs use certain pre-training tasks~\cite{hu2019strategies} to learn the graph's general representation under the unsupervised setting. In order to perform graph classification tasks, a part of label data will be used to fine-tuning the models~\cite{qiu2020gcc}.\\
\textbf{Contrastive Learning} \ Contrastive learning has been widely used for unsupervised learning by training an encoder that can capture similarity from data. The contrastive loss is normally a scoring function that increases the score on the single matched instance and decreases the score on multiple unmatched instances~\cite{oord2018representation,wu2018unsupervised}. In the graph domain, DGI~\cite{velivckovic2018deep} is the first GNN model utilizing the idea of contrastive learning, where the mutual information between nodes and the graph representation is defined as the contrastive metric. HDGI~\cite{ren2019heterogeneous} extends the mechanism to heterogeneous graphs. InfoGraph~\cite{sun2019infograph} performs contrastive learning in semi-supervised graph-level representation learning. When faced with the task of supervised learning, such as graph classification, we also need to use the advantage of contrastive learning to capture similarity. GCC~\cite{qiu2020gcc} utilizes contrastive learning to pre-train a model that can serve for the downstream graph classification task by fine-tuning. Compared to them, our method is an end-to-end model and performs label contrastive coding to encourage instance-level intra-class compactness and inter-class separability. 

\vspace{-15pt}
\section{Proposed Method}\label{sec:method} 
\vspace{-10pt}
In this section, we introduce the label contrastive coding based graph neural network ({\our}). Before introducing {\our}, we provide the preliminaries about graph classification first.
\vspace{-10pt}
\subsection{Preliminaries}

The goal of graph classification is to predict class labels of graphs based on the graph structural information and node contents. Formally, we denote it as follows:
\subsubsection{Graph Classification} Given a set of labeled graphs $\mathbb{G}_L = \{(\mathcal{G}_1,y_1),(\mathcal{G}_2,y_2),\dots\}$ and $y_i \in \mathbb{Y}$ is the corresponding label of $\mathcal{G}_i$. The task is to learn a classification function $\mathit{f}:\mathcal{G}\longrightarrow\mathbb{Y}$ to make predictions for unseen graphs $\mathbb{G}_U$.

\vspace{-10pt}
\subsection{{\our} Architecture Overview}
A learning process illustration of the proposed {\our} is shown in Figure~\ref{fig:framework}. Usually, for the input graph, we need to extract the latent features that can serve the graph classification through a high-performance graph encoder. In order to cooperate with the proposed mixed loss (\textit{Label Contrastive Loss} \& \textit{Classification Loss}), {\our} contains two graph encoder $f_k$ and $f_q$, which serve for encoding input key graphs and query graphs respectively. \textit{Label Contrastive Loss} encourages instance-level intra-class compactness and inter-class separability simultaneously by keeping intermediate discriminative representations, while \textit{Classification Loss} ensures the class-level discriminability.
A dynamic memory bank containing key graph representations and corresponding labels works for label contrastive loss calculation. A graph classifier takes the representations from the graph encoder $f_q$ as its input to predict the graph labels. In the following parts, we will elaborate on each component and the learning process of {\our} in detail.

\vspace{-10pt}
\subsection{Label Contrastive Coding}
Existing contrastive learning has been proved a success in training an encoder that can capture the universal structural information behind graph data~\cite{qiu2020gcc}. In the graph classification task, we focus on classification-related structural patterns compared with the universal structural patterns. Therefore, our proposed label contrastive coding learns to discriminate between instances with different class labels instead of treating each instance as a distinct class of itself and contrasting to other distinct classes.

Contrastive learning can be considered as learning an encoder for a dictionary look-up task~\cite{he2020momentum}. We can describe the contrastive learning as follows. Given an encoded query $\textbf{q}$ and a dictionary containing $m$ encoded keys $\{\textbf{k}_1,\textbf{k}_2,\dots,\textbf{k}_m\}$, there is only a single positive key $\textbf{k}_+$ (normally encoded from the same instance as $\textbf{q}$). The loss of this contrastive learning is low when $\textbf{q}$ is similar to the positive key $\textbf{k}_+$ while dissimilar to negative keys for $\textbf{q}$ (all other keys in the dictionary). A widely used contrastive loss is InfoNCE~\cite{oord2018representation} like:

\begin{equation}
\mathcal{L} = -\log\frac{\mathrm{exp}(\textbf{q}\cdot\textbf{k}_+/\tau)}{\sum_{i=1}^{m}\mathrm{exp}(\textbf{q}\cdot\textbf{k}_i/\tau)}	
\end{equation}

Here, $\tau$ is the temperature hyper-parameter~\cite{wu2018unsupervised}. Essentially, the loss of InfoNCE is a classification loss aiming to classify $\textbf{q}$ from $m=1$ classes to the same class as $\textbf{k}_+$. 

However, when facing graph classification tasks, the class labels have been determined, and we hope to import known label information in the training data to assist contrastive learning to serve the graph classification task. In this way, we design the label contrastive coding.

\subsubsection{Define similar and dissimilar} 
In the graph classification task, we seek that instances with the same label can be pulled closer, while instances with different labels will be pushed away from each other. Therefore, in the label contrastive coding, we consider two instances with the same label as a similar pair while treating the pair consisting of different label instances as dissimilar. 

\begin{figure}[t]
	\centering
	\begin{minipage}[l]{1\columnwidth}
		\centering
		\includegraphics[width=0.8\textwidth]{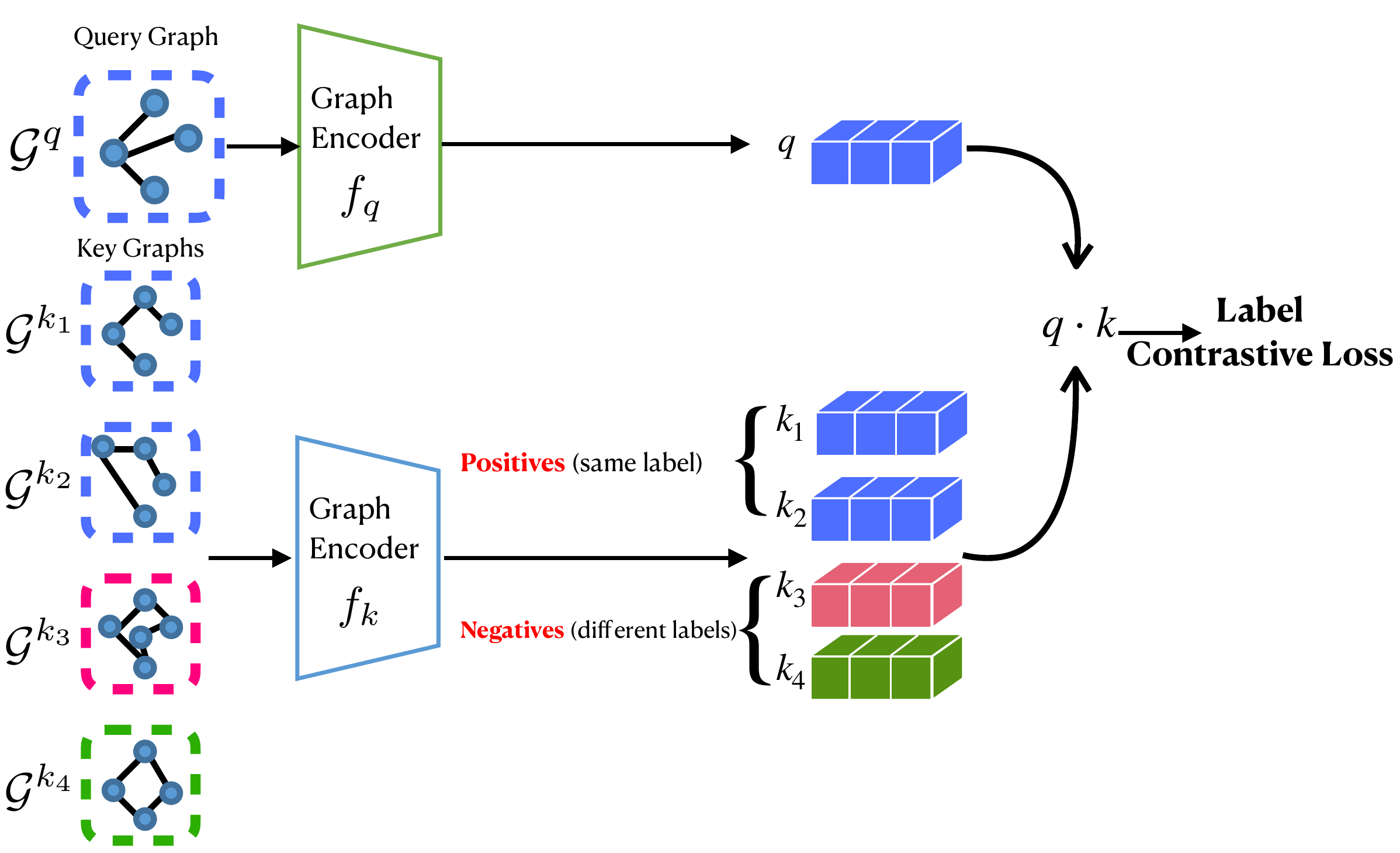}
	\end{minipage}
	\vspace{-5pt}
	\caption{\small \textit{Label Contrastive Loss}. The query graph $\mathcal{G}^q$ and key graphs $\mathcal{G}^{k}$ are encoded by $f_q$ and $f_k$ to low-dimensional representations $\textbf{q}$ and $\textbf{k}$ respectively. $\textbf{k}_1$ and $\textbf{k}_2$ having the same label as $\textbf{q}$ are denoted as positive keys. $\textbf{k}_3$ and $\textbf{k}_4$ are negative keys due to different labels. The label contrastive loss encourage the model to distinguish the similar pair $(\mathcal{G}^q , \mathcal{G}^{k_1})$ and $(\mathcal{G}^q , \mathcal{G}^{k_2})$ from dissimilar instance pairs, e.g., $(\mathcal{G}^q , \mathcal{G}^{k_3})$. }
	\label{fig:label_loss}
	\vspace{-10pt}
\end{figure}
\vspace{-10pt}
\subsubsection{Label contrative loss}  
Still from a dictionary look-up perspective, given an labeled encoded query $(\textbf{q},y)$, and a dictionary of $m$ encoded labeled keys $\{(\textbf{k}_1,y_1)$,\\$(\textbf{k}_2,y_2),\dots,(\textbf{k}_m,y_m)\}$, the 
positive keys $\textbf{k}_+$ in label contrastive coding are the keys $\textbf{k}_i$ where $y_i = y$. The label contrastive coding looks up the positive keys $\textbf{k}_i$ that the query $\textbf{q}$ matches in the dictionary. For the encoded query $(\textbf{q},y)$, its label contrastive loss $\mathcal{L}_{LC}$ is calculated by

\begin{equation}\label{equ:lc}
\mathcal{L}_{LC}(\textbf{q},y) = -\log \frac{\sum_{i=1}^{m}\mathbbm{1}_{y_i = y}\cdot\exp(\textbf{q}\cdot \textbf{k}_i/\tau)}{\sum_{i=1}^{m}\exp (\textbf{q}\cdot \textbf{k}_i/\tau)}
\end{equation}
Here,$\mathbbm{1}_{statement} \in \{0, 1\}$ is a binary indicator that returns $1$ if the statement is true.
We illustrate the label contrastive loss in Figure~\ref{fig:label_loss} for reference. In {\our}, key graph representations are stored in a dynamic memory bank. For the sake of brevity, we have not shown in Figure~\ref{fig:label_loss}. We introduce the dynamic memory bank and the updating process next.
\vspace{-10pt}
\subsubsection{The dynamic memory bank}
In label contrastive coding, the $m$-size dictionary is necessary. We use a dynamic memory bank to work as a dictionary. In order to fully utilize label information, the size of the memory bank is equal to the size of the set of labeled graphs $\mathbb{G}_L$, i.e., $m = |\mathbb{G}_L|$. The memory bank contains both the encoded low-dimensional key graph representations along with the corresponding labels, i.e., $\{(\textbf{k}_1,y_1),(\textbf{k}_2,y_2),\dots,(\textbf{k}_{|\mathbb{G}_L|},y_{|\mathbb{G}_L|})\}$. Based on the conclusions in MoCo~\cite{he2020momentum}, the key graph representations should be kept as consistent as possible when the graph encoder $f_k$ encoder evolves during training. Therefore, in each training epoch, newly encoded key graphs will dynamically replace the old version in the memory bank.

\vspace{-10pt}
\subsection{Graph Encoder Design}
For given graphs $\mathcal{G}^q$ and $\mathcal{G}^k$, {\our} empolys two graph encoders $f_q$ and $f_k$ to encode them to low-dimensional representations. 
\begin{equation}\label{equ:encode}
\begin{aligned}
\textbf{q} &= f_q(\mathcal{G}^q)\\\textbf{k} &= f_k(\mathcal{G}^k)
\end{aligned}
\end{equation}
In {\our}, $f_q$ and $f_k$ have the same structure. Graph neural network has proven its powerful ability to encode graph structure data~\cite{wu2020comprehensive}. Many potential graph neural networks can work as the graph encoder in {\our}. 

Two kinds of encoders are considered in {\our}. The first is Graph Isomorphism Network (GIN)~\cite{xu2018powerful}. GIN uses multi-layer perceptrons (MLPs) to conceive aggregation scheme and updates node representations as:

\begin{small}
	\begin{equation}
	{h}_v^{k} = \mathrm{MLP}^{(k)}\bigg((1+\epsilon^{(k)})+\sum\nolimits_{u\in\mathcal{N}(v)}{h}_u^{(k-1)}\bigg)
	\end{equation}
\end{small}
where $\epsilon$ is a learnable parameter or a fixed scalar, and $k$ represents $k$-th layer. Given
embeddings of individual nodes, the readout function is proposed by GIN to produce the representation $\boldsymbol{g}$ of the entire graph $\mathcal{G}$ for graph classification tasks:

\begin{small}
	\begin{equation}
	\boldsymbol{g} = \concatenate_{k=1}^K\bigg(\mathrm{SUM}(\{{h}_v^{k}| v\in \mathcal{G}\})\bigg)
	\end{equation}
\end{small}
Here, $\concatenate$ is the concatenation operator.

The second encoder we consider is Hierarchical Graph Pooling with
Structure Learning (HGP-SL)~\cite{zhang2019hierarchical}. HGP-SL incorporates graph pooling and structure learning into a unified module to generate hierarchical representations of graphs. HGP-SL proposes a graph pooling operation to identify a subset of informative nodes to form a new but smaller graph. The details about the Manhattan distance-based pooling operation can be referenced to \cite{zhang2019hierarchical}. For graph $\mathcal{G}$, HGP-SL repeats the graph convolution and pooling operations for several times and achieves multiple subgraphs in different layers: $\textbf{H}^1, \textbf{H}^2, \dots, \textbf{H}^K$. HGP-SL uses the concatenation of mean-pooling and max-pooling to aggregate all the node representations in the subgraph as follows:

\begin{small}
	\begin{equation}
	\textbf{r}^k = \mathcal{R}(\textbf{H}^k) = \sigma\bigg(\frac{1}{n^k}\sum_{p=1}^{n^k}\textbf{H}^k(p,:) \concatenate\max_{q=1}^d\textbf{H}^k(:,1)\bigg)
	\end{equation}
\end{small}

where $\sigma$ is a nonlinear activation function. $n^k$ is the node number in the $k$-th layer subgraph. In order to achieve the final representation $\boldsymbol{g}$ of the entire graph $\mathcal{G}$, another readout function is utilized to combine subgraphs in different layers.

\begin{small}
	\begin{equation}
	\boldsymbol{g} = \mathrm{SUM}(\textbf{r}^k | k =  1, 2, \dots, K)
	\end{equation}
\end{small}

In the experiment section, we will show the performance along with the analysis of using GIN and HGP-SL as graph encoders in {\our}.

\vspace{-10pt}
\subsection{{\our} Learning}\label{sec:learn}
The training process illustration is provided in Figure~\ref{fig:framework}. During the training process, the input of {\our} is a batch of labeled graphs $\mathbb{G}_b \subset \mathbb{G}_L$. For each mini-batch iteration, the set of key graphs and the set of query graphs are the same as $\mathbb{G}_b$. The graph encoder $f_q$ and $f_k$ will be initialized with the same parameters ($\theta_q = \theta_k$). The memory bank's size is equal to the size of the set of labeled graphs $\mathbb{G}_L$. The labeled graph $\mathcal{G}^i$ with the label $y_i$ is assigned with a random representation to initialize the memory bank. 
The set of key graphs will be encoded by $f_k$ to low-dimensional key graph representations $\mathbb{K}$, which will replace the corresponding representations in the memory bank. The set of query graphs are encoded by $f_q$ to query graph representations $\mathbb{Q}$, whereas $\mathbb{Q}$ is also the input of the graph classifier. In {\our}, a logistic regression layer serves as the graph classifier. Based on the output of the graph classifier, \textit{Classification Loss} can be calculated by:

\begin{equation}
\mathcal{L}_{Cla} = - \frac{1}{|\mathbb{Q}|}\sum_{\textbf{q}_i \in \mathbb{Q}} \sum_{j \in \mathbb{Y}} \mathbbm{1}_{\textbf{q}_i,j} \textup{log}(p_{\textbf{q}_i,j})
\label{eq:loss}
\end{equation}
where $\mathbbm{1}$ is a binary indicator (0 or 1) that indicates whether label $j$ is the correct classification for the encoded query graph $\textbf{q}_i$. Besides, $p_{\textbf{q}_i,j}$ is the predicted probability.

$\mathbb{Q}$ and the memory bank work together to implement the label contrastive coding described in previous parts. Based on the Equ.~\ref{equ:lc}, \textit{Label Contrastive Loss} of the mini-batch $\mathbb{G}_b$ is:

\begin{equation}\label{equ:lcb}
\mathcal{L}_{LC} = - \frac{1}{|\mathbb{Q}|}\sum_{\textbf{q}_i \in \mathbb{Q}}\mathcal{L}_{LC}(\textbf{q}_i,y_{\textbf{q}_i})
\end{equation}

In order to train the model by utilizing label information more effectively and comprehensively, we try to minimize the following mixed loss combining both the \textit{Label Contrastive Loss} and the \textit{Classification Loss}:

\begin{equation}\label{equ:mix}
\mathcal{L}_{total} = \mathcal{L}_{Cla} + \beta\  \mathcal{L}_{LC}
\end{equation}
Here, the hyper-parameter $\beta$ controls the relative weight between the label contrastive loss and the classification loss. The motivation behind $\mathcal{L}_{total}$ is that $\mathcal{L}_{LC}$ encourages instance-level intra-class compactness and inter-class separability while
$\mathcal{L}_{Cla}$ ensures the discriminability of classes. The graph encoder $f_q$, and the graph classifier can be updated end-to-end by back-propagation according to the loss $\mathcal{L}_{total}$. The parameters $\theta_k$ of $f_k$ follows a momentum-based update mechanism as MoCo~\cite{he2020momentum} instead of the back-propagation way. Specifically, the momentum updating process is:
\begin{equation}\label{equ:momentum}
\theta_k \longleftarrow \alpha\theta_k + (1-\alpha)\theta_q
\end{equation}
where $\alpha \in [0,1)$ is the momentum weight to control the speed of $f_k$ evolving. We use this momentum-based update mechanism not only to reduce the overhead of backpropagation but also to keep the key graph representations in the memory bank as consistent as possible despite the encoder's evolution.

After completing the model training, the learned graph encoder $f_q$ along with the graph classifier can be used to perform graph classification tasks for the unlabeled graphs $\mathbb{G}_U$.

\vspace{-10pt}
\section{Experiments}\label{sec:experiment}
\vspace{-10pt}
\subsection{Experiment Settings}
\vspace{-5pt}
\subsubsection{Datasets}

We test our algorithms on 8 widely used datasets. Three of them are social networks benchmark datasets: IMDB-B, IMDB-M, and COLLAB; the rest five datasets: MUTAG, PROTEINS, PTC, NCI1, and D$\&$D, belong to biological graphs datasets~\cite{yanardag2015deep,xinyi2018capsule,xu2018powerful}. Each dataset of these contains many graphs, and each graph is assigned with a label. The statistics of these datasets are summarized in Table~\ref{tab:dataset_stat}. What should be mentioned is that the biological graphs have categorical node attributes, while social graphs do not come with node attributes. In this paper, for the encoders requiring node attributes as input, we follow~\cite{xu2018powerful} to use one-hot encodings of node degrees as the node attributes on datasets without node features.
\begin{table}[h]
	\vspace{-15pt}
	\caption{\small Datasets in the Experiments}
	\vspace{-10pt}
	\renewcommand\arraystretch{1}
	\centering
	\begin{threeparttable}
		\begin{tabular}{c | c c c c }
			\hline 
			Datasets&$\#$ graphs&Avg $\#$ nodes&Avg $\#$ edges&$\#$ classes\\
			\hline
			IMDB-B&$1000$&$19.77$&$96.53$&$2$\\
			IMDB-M&$1500$&$13.00$&$65.94$&$3$\\
			COLLAB&$5000$&$74.49$&$2457.78$&$3$\\
			MUTAG&$188$&$17.93$&$19.79$&$2$\\
			PROTEINS&$1113$&$39.06$&$72.82$&$2$\\
			PTC&$344$&$25.56$&$25.56$&$2$\\
			NCI1&$4110$&$29.87$&$32.30$&$2$\\
			D$\&$D&$1178$&$284.32$&$715.66$&$2$\\
			\hline
		\end{tabular}
		
	\end{threeparttable}
	\vspace{-30pt}
	\label{tab:dataset_stat}
\end{table}

\subsubsection{Methods Compared}
We select three categories of models as our comparison methods:\\
\noindent
\vspace{-12pt}
\begin{itemize}	
	\item~\underline{Kernel-based method}: Weisfeiler-Lehman subtree kernel (\textbf{WL})~\cite{shervashidze2011weisfeiler}, \textbf{AWE}~\cite{ivanov2018anonymous}, and Deep Graph Kernel (\textbf{DGK})~\cite{yanardag2015deep}: They first decompose graphs into sub-components based on the kernel definition, then learn graph embeddings in a feature-based manner. For graph classification tasks, a machine learning model (i.e., SVM) will be used to perform the classification with learned graph embeddings.
	\item~\underline{Graph embedding-based methods}: \textbf{Sub2vec}~\cite{adhikari2018sub2vec}, \textbf{Graph2vec}~\cite{narayanan2017graph2vec}: They extend document embedding neural networks to learn representations of entire graphs. A machine learning model (i.e., SVM) work on the classification tasks with learned graph representations.
	\item~\underline{Graph neural network methods}: \textbf{GraphSAGE}~\cite{hamilton2017inductive}, \textbf{GCN}~\cite{kipf2016semi}, \textbf{DCNN}~\cite{atwood2016diffusion}: They are designed	to learn meaningful node level representations. A readout function is empolyed to summarize the node representations to the graph representation for graph-level classification tasks;  \textbf{DGCNN}~\cite{zhang2018end}, \textbf{CapsGNN}~\cite{xinyi2018capsule}, \textbf{HGP-SL}~\cite{zhang2019hierarchical}, \textbf{GIN}~\cite{xu2018powerful}, \textbf{InfoGraph}~\cite{sun2019infograph}: They are GNN-based algorithms with the pooling operator for graph representation learning. Then a classification layer will work as the last layer to implement graph classification;
	\textbf{GCC}~\cite{qiu2020gcc}: It follows pre-training and fine-tuning paradigm for graph representation learning. A linear classifier is used to support the fine-tuning targeing graph classification;
	$\textbf{{\our}}_\textbf{GIN}$, $\textbf{{\our}}_\textbf{HGP-SL}$: They are two variants of the proposed {\our}. $\our_{GIN}$ uses GIN~\cite{xu2018powerful} as the graph encoders, and $\our_{HGP-SL}$ sets the graph encoders as HGP-SL~\cite{zhang2019hierarchical}. 
\end{itemize}

\vspace{-10pt}
\subsubsection{Experiment Configurations}
\vspace{-10pt}
We adopt two graph model structures: GIN~\cite{xu2018powerful} and HGP-SL~\cite{zhang2019hierarchical} as the graph encoders. For {\our} with different encoders, we follow the model configurations from the initial papers as the default settings. For the {\our} structure, we choose the hidden representation dimension as 64 and 128 for two respective encoders; the contrastive loss weight $\beta\in \{0.1, 0.6,\dots, 1.0\}$; the momentum term $\alpha\in[0.0, 1.0)$; the temperature $\tau = 0.07$. For the graph classification tasks, to evaluate the proposed {\our} we adopt the procedure in~\cite{xu2018powerful,xinyi2018capsule} to perform 10-fold cross-validation on the aforementioned datasets.  For the training process of {\our}, we select the Adam~\cite{adam} as the optimizer, and tune the hyperparameters for each dataset as: (1) the batch size $\in\{32, 128, 512\}$;  (2) the learning rate $\in\{0.01, 0.001\}$; (3) the dropout rate $\in\{0.0, 0.5\}$; (4) number of training epochs 1000 and select the epoch as the same with~\cite{xu2018powerful}. We run the experiments on the Server with 3 GTX- 1080 ti GPUs, and all codes are implemented in Python3. Code and supplementary materials are available at:
\textit{Anonymous link}\footnote{https://www.dropbox.com/sh/kc7xf42kz4lqx9a/AAC9wKim768TBNocN1JNPudFa?dl=0}. 
\vspace{-10pt}
\subsection{Experimental Results and Analysis}
\vspace{-10pt}
\subsubsection{Overall evaluation}
We present the main experimental results in Table~\ref{tab:results_all}.  For the graph datasets that comparison methods do not have the results in the original papers, we denote it as "$-$". From the table, we can observe that {\our} outperforms all comparison methods on 7 out of the total 8 datasets. The improvement is especially evident on datasets such as IMBD-B and D$\&$D, which can be up to about $1.0\%$. At the same time, we can find that {\our} using different graph encoders have achieved SOTA performance on different datasets ($\our_{GIN}$ in 3 datasets; $\our_{HGP-SL}$ in 4 datasets). The results also show that for different data sets, the selection of graph encoders has a critical impact on performance Nonetheless {\our} generally outperforms all other baselines methods.

\begin{adjustbox}{angle=90}
	
	\renewcommand\arraystretch{1.5}
	\centering
	\begin{threeparttable}
		\caption{\small Test Sets Classification Accuracy on All Datasets. We use \textbf{bold} to denote the best result on each dataset.}
		\begin{tabular}{p{2cm} p{2cm} c c c c c c c c}
			\toprule[1.5pt]
			%
			\textbf{Categories}&\textbf{Methods}&IMDB-B&IMDB-M&COLLAB&MUTAG&PROTEINS&PTC&NCI1&D$\&$D\\
			\hline
			\multirow{3}*{Kernels}&WL&$73.4\pm 4.6$&$49.3\pm 4.8$&$79.0\pm 1.8$&$82.1 \pm 0.4$&$76.2\pm 4.0$&$-$&$76.7\pm 2.0$&$76.4 \pm 2.4$\\
			&AWE&$74.5\pm 5.9$&$51.5\pm 3.6$&$73.9\pm 1.9$&$87.9 \pm 9.8$&$-$&$-$&$-$&$71.5 \pm 4.0$\\
			&DGK&$67.0\pm 0.6$&$44.6\pm 0.5$&$73.1\pm 0.3$&$87.4 \pm 2.7$&$75.7\pm 0.5$&$60.1\pm 2.5$&$80.3\pm 0.5$&$73.5\pm 1.0$\\
			\hdashline
			Graph&Graph2vec&$71.1\pm 0.5$&$50.4\pm 0.9$&$-$&$83.2 \pm 9.3$&$73.3\pm1.8$&$60.2\pm6.9$&$73.2\pm 1.8$&$-$\\
			Embedding&Sub2vec&$55.2\pm 1.5$&$36.7\pm 0.8$&$-$&$61.0 \pm 15.8$&$-$&$60.0\pm 6.4$&$-$&$-$\\
			\hdashline
			\multirow{7}*{GNNs}&DCNN&$72.4\pm 3.6$&$49.9\pm 5.0$&$79.7\pm 1.7$&$79.8 \pm 13.9$&$65.9\pm 2.7$&$-$&$74.7\pm 1.3$&$-$\\
			&GCN&$73.3\pm 5.3$&$51.2\pm 5.1$&$\mb{80.1\pm 1.9}$&$87.2\pm 5.1$&$75.2\pm 3.6$&$-$&$76.3\pm 1.8$&$73.3\pm 4.5$\\
			&GraphSAGE&$72.4\pm 3.6$&$49.9\pm 5.0$&$79.7\pm 1.7$&$79.8 \pm 13.9$&$65.9\pm 2.7$&$$&$74.7\pm 1.3$&$-$\\
			&DGCNN&$70.0\pm 0.9$&$47.8\pm 0.9$&$73.8\pm 0.5$&$85.8\pm 1.7$&$75.5\pm 0.9$&$58.6 \pm 2.5$&$74.4\pm 0.5$&$79.4\pm 0.9$\\
			&CapsGNN&$73.1\pm 4.8$&$50.3 \pm 2.7$&$79.6\pm 0.9$&$86.7\pm 6.9$&$76.3\pm 3.6$&$-$&$78.4\pm 1.6$&$75.4\pm 4.2$\\
			&HGP-SL&$-$&$-$&$-$&$ 82.2\pm 0.6$&$84.9\pm 1.6$&$-$&$78.5\pm 0.8$&$81.0\pm 1.3$\\
			&GIN&$75.1\pm 5.1$&$52.3\pm 2.8$&$80.2\pm 1.9$&$89.4\pm 5.6$&$76.2\pm 2.8$&$64.6\pm 7.0$&$82.7\pm 1.7$&$-$\\
			&InfoGraph&$73.0\pm 0.9$&$49.7\pm 0.5$&$-$&$89.0\pm 1.1$&$-$&$61.7\pm 1.4$&$-$&$-$\\
			&GCC&$73.8$&$50.3$&$81.1$&$-$&$-$&$-$&$-$&$-$\\
			\hline
			\multirow{2}*{Proposed}&${\our}_{GIN}$&$\mb{76.1\pm 6.9}$&$\mb{52.4\pm 6.7}$&$72.3\pm 6.3$&$89.9\pm 4.8$&$76.9\pm 6.8$&$64.7\pm 2.0$&$\mb{82.9\pm 3.6}$&$77.4\pm 1.2$\\
			&${\our}_{HGP-SL}$&$75.4\pm 1.5$&$46.5\pm 1.3$&$77.5\pm 1.2$&$\mb{90.5\pm 2.3}$&$\mb{85.2\pm 2.4}$&$\mb{65.9\pm 2.8}$&$78.8\pm 4.4$&$\mb{81.8\pm 3.6}$\\
			
			\bottomrule[1.5pt]
			\end{tabular} 
			\label{tab:results_all}
			\end{threeparttable}
			\vspace{-20pt}
			\end{adjustbox}
			
\begin{figure}[t]
			\vspace{-10pt}
			\centering
			\subfigure[IMDB-B]
			{
				\begin{minipage}[l]{0.4\columnwidth}
				\centering
				\includegraphics[width=1\textwidth]{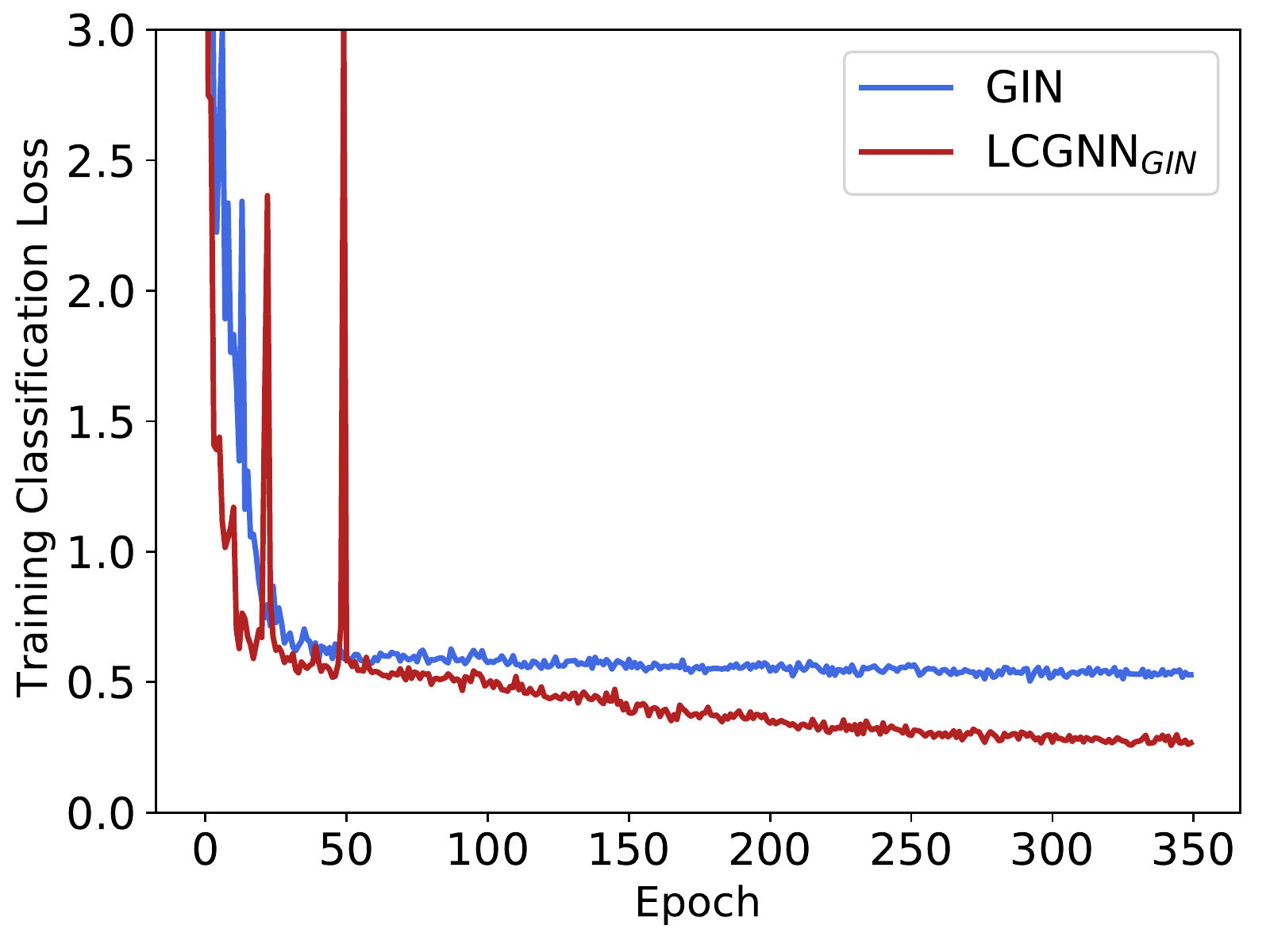}
				\end{minipage}
				\label{fig:orl_1}
			}
			\subfigure[IMDB-M]{
				\begin{minipage}[l]{0.4\columnwidth}
				\centering
				\includegraphics[width=1\textwidth]{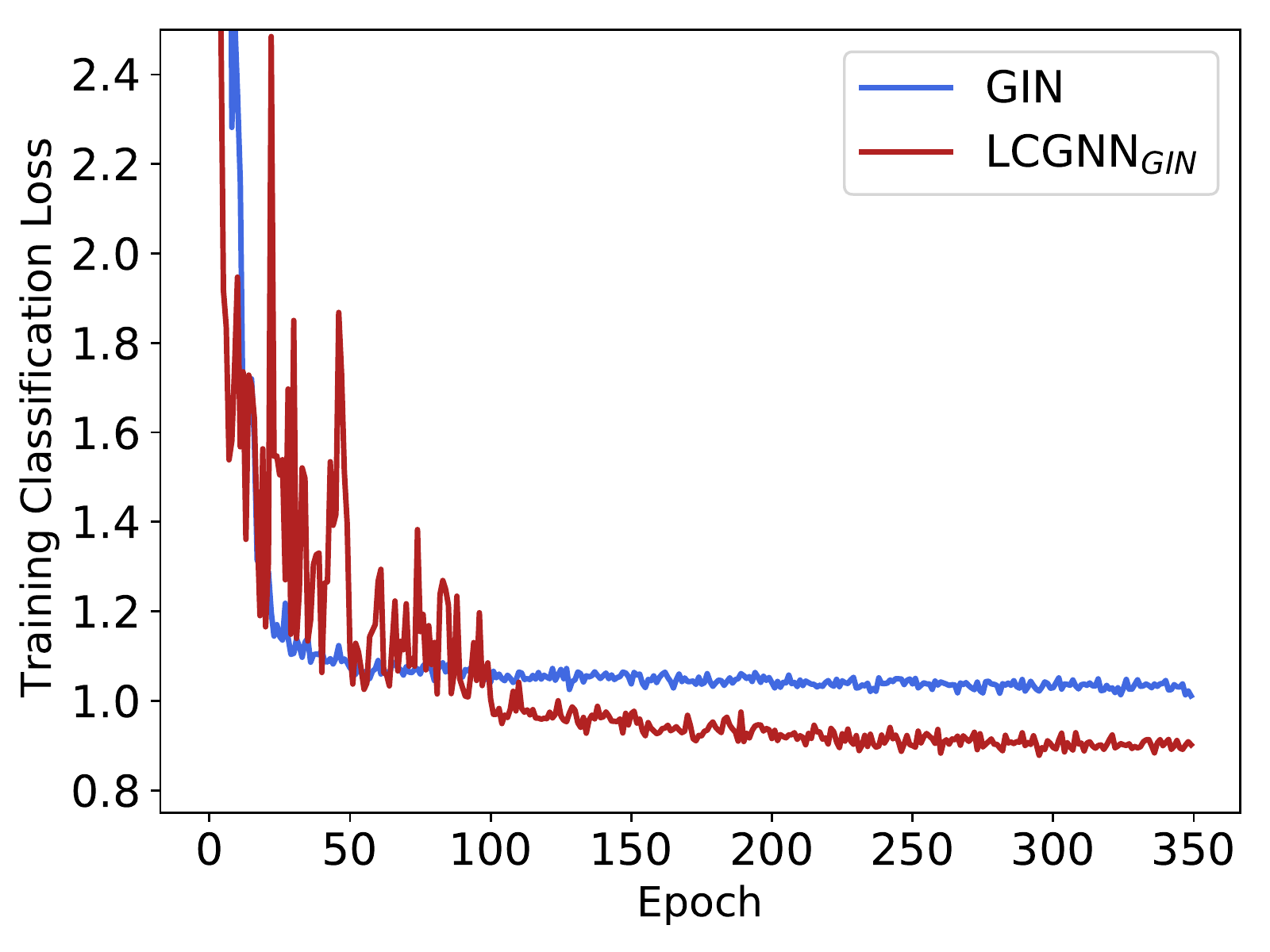}
				\end{minipage}
				\label{fig:orl_2}
			}
			\subfigure[MUTAG]{ \label{fig:mnist_train}
				\begin{minipage}[l]{0.4\columnwidth}
				\centering
				\includegraphics[width=1\textwidth]{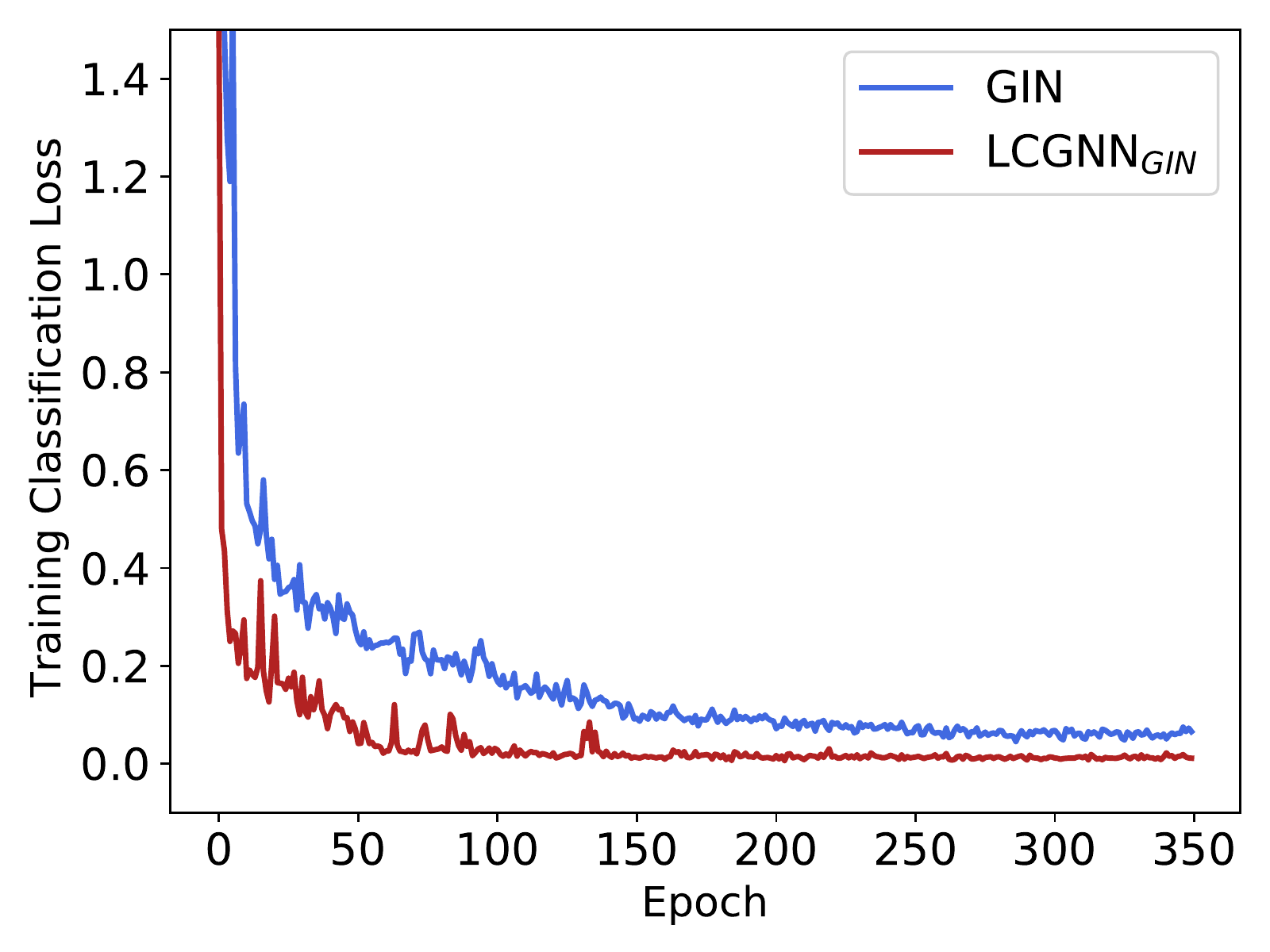}
				\end{minipage}
				\label{fig:dnn_1}
			}
			\subfigure[PROTEINS]{ \label{fig:mnist_test}
				\begin{minipage}[l]{0.4\columnwidth}
				\centering
				\includegraphics[width=1\textwidth]{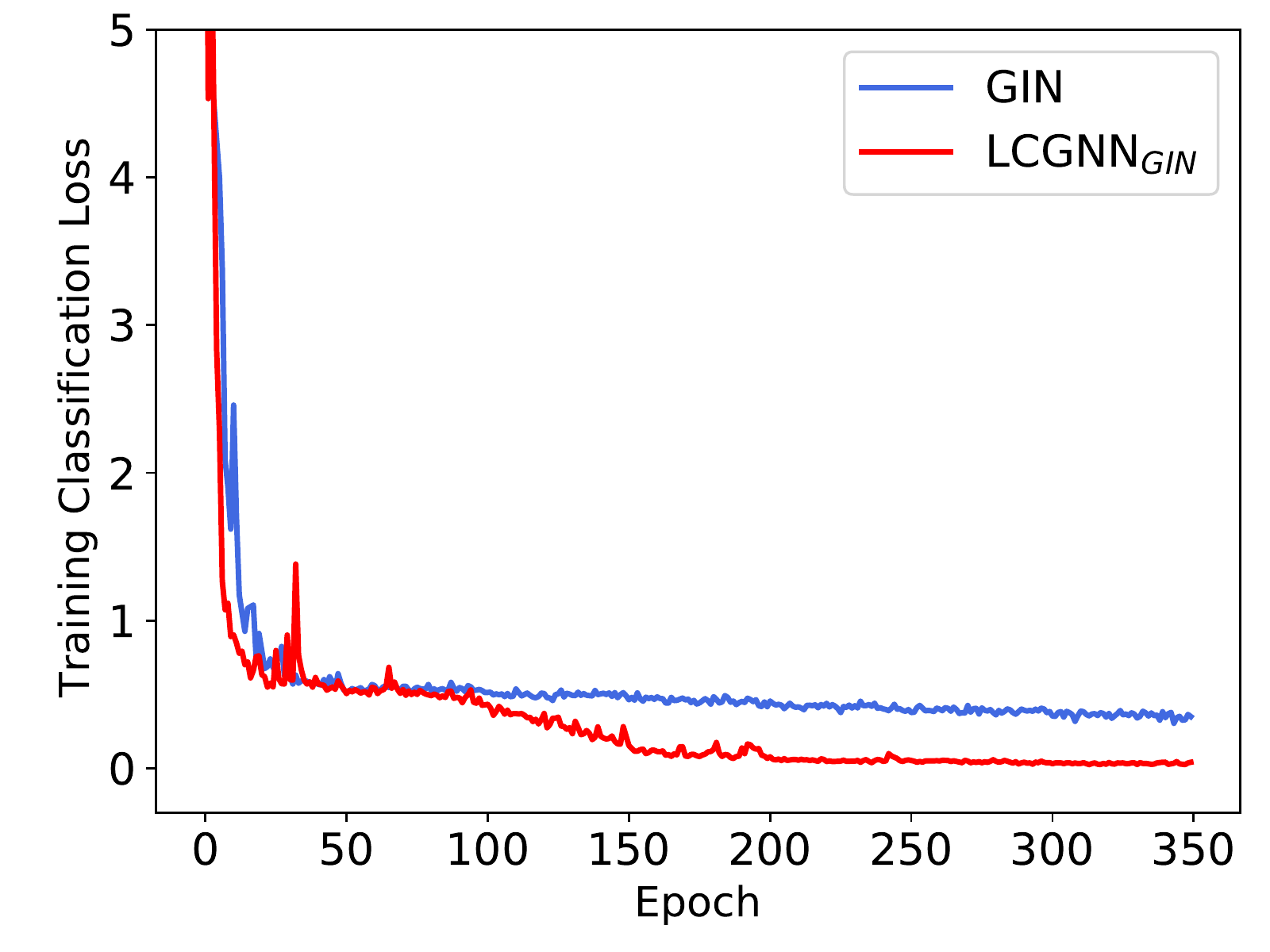}
				\end{minipage}
				\label{fig:dnn_2}
			}
			\vspace{-10pt}
			\caption{\small Training \textit{Classification Loss} versus Training Epoch}\label{fig:trainloss}
			\vspace{-20pt}
			\end{figure}

			We also note that, compared to the baseline methods GIN and HGP-SL, $\our_{GIN}$ and $\our_{HGP-SL}$ can acquire better results when adopting them as corresponding encoders. For the COLLAB dataset results, {\our} actually achieves much higher performance compared with the result we get when running GIN source code ($71.7\pm 3.5$). However, the result reported by the original paper~\cite{xu2018powerful} is $80.1\pm 1.9$, which we also report in Table~\ref{tab:results_all}. To further evaluate the advantages of {\our} and highlight the effectiveness of \textit{Label Contrastive Loss}, we compare the classification loss during the training processes and show the curves of GIN and {\our} in Figure~\ref{fig:trainloss}. From the figure, we can see that not only {\our} has a faster convergence rate, but also can finally converge to lower classification loss. The classification loss comparison results on other datasets are also consistent, but we did not show them all due to the space limitation. Thus we can conclude that with the support of label contrastive coding, {\our} has better potential on graph classification tasks.  
			
			Besides, through the comparison between GCC and {\our}, we can find that for the task of graph classification, The proposed label contrastive coding shows more advantages than the contrastive coding in GCC. We believe that the contrastive coding in GCC mainly focuses on learning universal representations. The label contrastive coding in {\our} has a stronger orientation for representation learning, that is, extracting features that significantly affect on the intra-class compactness and inter-class separability.
			\begin{table}[h]
			\vspace{-15pt}
			\caption{\small Experiments with Less Labeled Training Data}
			\vspace{-5pt}
			\scriptsize
			\renewcommand\arraystretch{1.3}
			\centering
			\begin{threeparttable}
			\begin{tabular}{c c c c c c c}
			\toprule[1.5pt]
			\multirow{2}*{Datasets}&\multirow{2}*{Methods}&\multicolumn{5}{c}{Training Ratio}\\
			
			\cline{3-7}
			&&60$\%$&70$\%$&80$\%$&90$\%$&100$\%$\\
			\hline
			\multirow{2}*{IMDB-B} &GIN&$61.8$&$65.4$&$69.2$&$70.5$&$75.1$\\
			&${\our}_{GIN}$&$\mb{66.3}$&$\mb{70.8}$&$\mb{71.3}$&$\mb{72.2}$&$\mb{76.1}$\\
			\hline
			\multirow{2}*{IMDB-M} &GIN&$40.5$&$41.4$&$41.8$&$46.0$&$52.3$\\
			&${\our}_{GIN}$&$\mb{43.4}$&$\mb{42.8}$&$\mb{43.6}$&$\mb{48.1}$&$\mb{52.4}$\\
			\bottomrule[1.5pt]
			\end{tabular}
			
			\end{threeparttable}
			\vspace{-25pt}
			\label{tab:less_labeled_data}
			\end{table}
			
			\vspace{-15pt}
			\subsubsection{Performance with less labeled data}
			To validate our claim that {\our} can utilize label information more comprehensively and use fewer label data to achieve comparative performance, we conduct experiments with less training data. For each fold of cross-validation, we extract only part of the training set (e.g., $60\%$ of the data in the training set) as the training data and maintain the test set as the same. We present the results in Table~\ref{tab:less_labeled_data}. In Table~\ref{tab:less_labeled_data}, the training ratio denotes how much data in the training set is extracted as the training data. When the training ratio is $100\%$, it means using the full training set in each fold. From the results, it is obvious that ${\our}_{GIN}$ can always outperform the baseline GIN when using less training data. What's more, in many cases when ${\our}_{GIN}$ with less training data (e.g., $70\%$ training data for ${\our}_{GIN}$ while $80\%$ for GIN; $60\%$ for ${\our}_{GIN}$ while $70\%$ for GIN), ${\our}_{GIN}$ still obtains more competitive results than GIN. The experimental results demonstrate that {\our} can utilize the same amount of training data more comprehensively and efficiently. The capability also makes {\our} possible to learn with less training data to obtain a better performance than comparison methods when they need more training data.
			\begin{figure}[t]
			\centering
			\vspace{-16pt}
			\begin{minipage}[l]{1\columnwidth}
			\centering
			\includegraphics[width=0.8\textwidth]{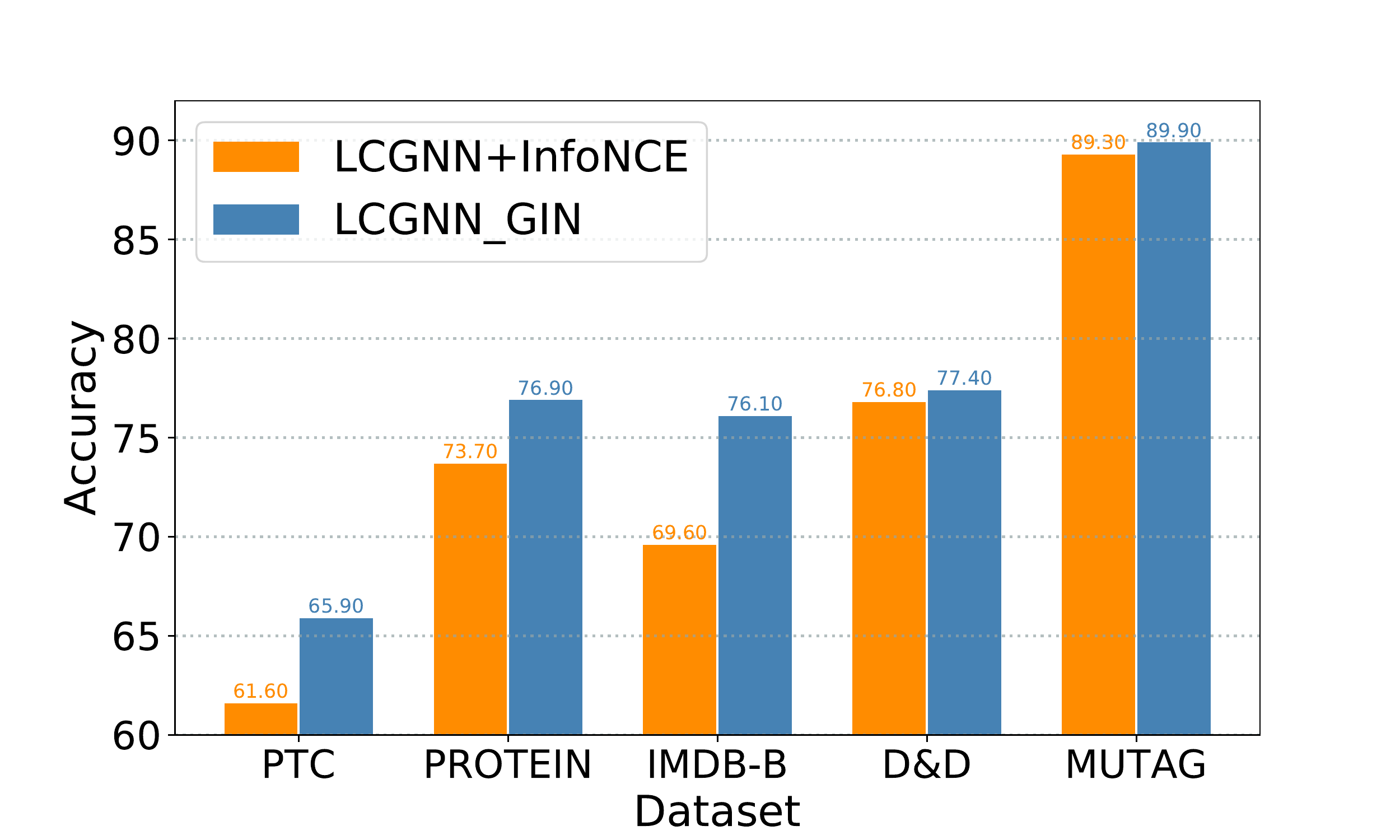}
			\end{minipage}
			\vspace{-10pt}
			\caption{\small The effectiveness of \textit{Label Contrastive Loss}}
			\label{fig:no_label_contrastive}
			\vspace{-15pt}
			\end{figure}
			
			\vspace{-15pt}
			\subsubsection{The effectiveness of the label contrastive coding}
			In order to further verify the effectiveness of the proposed label contrastive coding on the task of graph classification, we conduct comparison experiments between ${\our}_{GIN}$ and ${\our}$+InfoNCE. Here, ${\our}$+InfoNCE replaces the label contrastive loss in ${\our}_{GIN}$ with InfoNCE loss~\cite{oord2018representation} but keeps other parts the same. We present the results in Figure~\ref{fig:no_label_contrastive}. The experimental results show that the performance of ${\our}_{GIN}$ on all data sets exceeds ${\our}$+InfoNCE, which also demonstrates that the label contrastive coding can effectively utilize label information to improve model performance. In addition, we observe that the performance of ${\our}$+InfoNCE is even worse than GIN. It verifies that the inherent large intra-class variations may import noise to graph classification tasks if we treat the intra-class instances as distinct classes like the existing comparative learning. 
			\begin{table}[h]
			\vspace{-20pt}
			\caption{\small $\text{{\our}}_{GIN}$ with Different Contrastive Loss Weight $\beta$}
			\vspace{-10pt}
			\scriptsize
			\renewcommand\arraystretch{1.5}
			\centering
			\begin{threeparttable}
			\begin{tabular}{c c c c c c c c c}
			\toprule[1.5pt]
			\multirow{2}*{Datasets}&\multicolumn{8}{c}{Contrastive Loss Weight $\beta$}\\
			
			\cline{2-9}
			&0.3&0.4&0.5&0.6&0.7&0.8&0.9&1.0\\
			\hline
			IMDB-B&$73.8$&$75.1$&$\textbf{76.1}$&$75.5$&$76.0$&$75.4$&$75.7$&$75.7$\\
			IMDB-M&$50.5$&$51.2$&$\textbf{52.4}$&$51.9$&$51.7$&$51.5$&$51.5$&$51.6$\\
			\bottomrule[1.5pt]
			\end{tabular}
			
			\end{threeparttable}
			\vspace{-25pt}
			\label{tab:different_contraloss_weight}
			\end{table}
			
			\vspace{-18pt}
			\subsubsection{Hyper-parameter $\beta$ analysis}
			We consider the influence of label contrastive loss weight term $\beta$ and conduct experiments with different values. The results is exhibited in Table~\ref{tab:different_contraloss_weight}. We select $\beta$ from $\{0.1,0.2,\dots, 1.0\}$, and find the trend of using a relatively larger $\beta$ inducing better results. Thus in the experiment, we empirically select from $\beta\in \{0.5, 0.6, \dots, 1.0\}$ to achieve the best performance. Nevertheless, we also observed that when $\beta$ gradually increases, the performance does not continue to increase. Our analysis is that when the label contrastive loss weight is too high, the learning of the model places too much emphasis on instance-level contrast. More fine-grained discriminative features on the instance-level will reduce the generalization performance of the model on the test set.
			
			\begin{figure}[h]
			\centering
			\vspace{-15pt}
			\begin{minipage}[l]{1\columnwidth}
			\centering
			\includegraphics[width=0.68\textwidth]{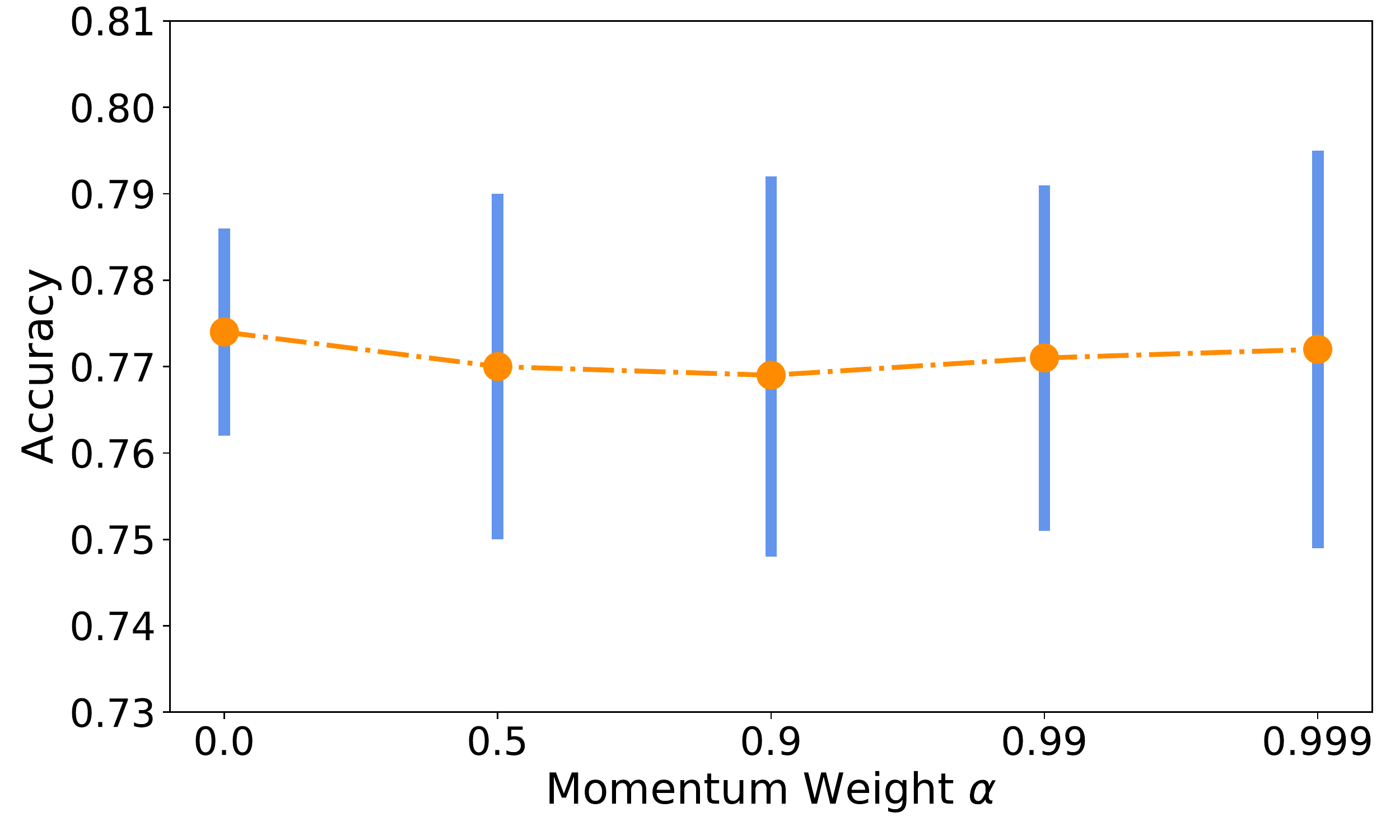}
			\end{minipage}
			\vspace{-10pt}
			\caption{{\our} with Different Momentum Weight}
			\label{fig:different_momentum_weight}
			\vspace{-20pt}
			\end{figure}

			\vspace{-15pt}
			\subsubsection{Momentum ablation} 
			The momentum term plays an important role in contrastive learning problems. In our experiments, we also try different momentum weight $\alpha$ when running $\our_{GIN}$ on D\&D and show the results in Figure~\ref{fig:different_momentum_weight}. Unlike~\cite{he2020momentum}, $\our_{GIN}$ also achieves good performance when $\alpha=0$. The main reason should be that the D\&D is not extremely large, which makes it easy for representations to ensure consistency during encoder evolving. Furthermore, in this set of experiments, the momentum term did not show much impact on Accuracy, that is, the model performance is relatively stable, which should be caused by the moderate-sized dataset as well.


\vspace{-15pt}
\section{Conclusion}\label{sec:conclusion}
\vspace{-12pt}
In this paper, we have introduced a novel label contrastive coding based graph neural network, {\our}, which works on graph classification tasks. We extend the existing contrastive learning to the supervised setting and define the label contrastive coding. The label contrastive coding treats instances with the same label as multiple positive instances, which is different from the single positive instance in unsupervised contrastive learning. The label contrastive coding can pull the same label instances closer and push the instances with different labels away from each other. We demonstrate the effectiveness of {\our} on graph classification tasks over eight benchmark graph datasets. The experimental results show that {\our} achieves SOTA performance in 7 datasets. Besides, {\our} can take advantage of label information more comprehensively. {\our} outperforms the baseline method when using less training data, which verifies this advantage.


\vspace{-10pt}
\bibliographystyle{splncs04}
\bibliography{reference}

\end{document}